\pdfoutput=1
\documentclass[11pt]{article}

\usepackage[]{emnlp2021}

\usepackage{times}
\usepackage{latexsym}
\usepackage{graphicx}
\usepackage{xcolor}
\usepackage{adjustbox}

\usepackage{booktabs}
\usepackage{tabularx}
\usepackage{csquotes}

\usepackage[T1]{fontenc}
\usepackage[utf8]{inputenc}

\usepackage{microtype}

\title{EmpBot: A T5-based Empathetic Chatbot focusing on Sentiments }

\author{Emmanouil Zaranis \\
  deeplab.ai \\
  National Technical University of Athens \\
  \texttt{m.zaranis@deeplab.ai} \\\And
Georgios Paraskevopoulos \\
  ATHENA R.C. / Athens,Greece \\
  Behavioral Signals / Athens,Greece\\
  \texttt{geopar@central.ntua.gr} \\\AND
  Athanasios Katsamanis \\
  ATHENA R.C. / Athens,Greece \\
  Behavioral Signals / Athens,Greece\\
  \texttt{nkatsam@athenarc.gr} \\\And
  Alexandros Potamianos \\
  ATHENA R.C. / Athens,Greece \\
  Behavioral Signals / Athens,Greece\\
  \texttt{potam@central.ntua.gr} \\}

\begin{document}
\maketitle
\begin{abstract}
In this paper, we introduce EmpBot: an end-to-end empathetic chatbot. Empathetic conversational agents should not only understand what is being discussed, but also  acknowledge the implied feelings of the conversation partner and respond appropriately. To this end, we propose a method based on a transformer pretrained language model (T5). Specifically, during finetuning we propose to use three objectives: response language modeling, sentiment understanding, and empathy forcing. The first objective is crucial for generating relevant and coherent responses, while the next ones are significant for acknowledging the sentimental state of the conversational partner and for favoring empathetic responses.  We evaluate our model on the EmpatheticDialogues dataset using both automated metrics and human evaluation. The inclusion of the sentiment understanding and empathy forcing auxiliary losses favor empathetic responses, as human evaluation results indicate, comparing with the current state-of-the-art.
\end{abstract}

\section{Introduction}\label{sec:RelatedWork}

Since dialogue is regarded as a fundamental and complex element of human cognition \cite{jurafskydialogsystems}, the development of systems capable of understanding human language and communicating with humans can have a significant impact. However, human communication requires the acknowledgment and the exchange of conversational partner's emotions, as emotions play an important role in developing a confidential relationship between the speaker and the listener.
\\Open domain conversational agents have been widely studied in the past years and both retrieval-based and generation-based approaches \citep{wu_2018_context_aware_prototype_editing,cai_2018_retrieval_memory,weston_2018_retrieve_refine} have been developed. However, prior research has shown that most of those  conversational agents are unable to imitate dialogues between humans, as the produced responses are generic and short \citep{vinyals_conv_model,li_2016c_persona_conv_model}. Several efforts have been made to make the conversationa more engaging by keeping track of the conversational context \citep{sordoni-context-sensitive,serban-hierarchical-query,serban-end-to-end-hierarchical,serban-vhred} or by producing more diverse responses \cite{li_diversity_objective_function,li-reinforcement-dialogue}. Subsequently, a recent trend that was followed by various researchers \citep{li_2016c_persona_conv_model,zhang_2018_personachat,kulikov_2018_search_eval_dialog_modeling,joshi_2017_pers,zemlyanskiy_2018_aiming_dialogue,mazare_2018_training,Dinan_2019_convai2,madotto_2019_personalizing_meta_learning,hancock_2019_feed_yourself_chatbot,yavuz_2019_deepcopy,wolf-transfer-transfo-personachat} in order to make the responses more coherent and consistent through the dialogue, was to produce personalized responses by conditioning the generation on a persona profile. %However, these works did not take into account the feelings of the conversational partner.
\\
Apart from understanding what is being discussed, a conversational agent should also acknowledge the emotional state of the conversational partner, as it is a significant part of human communication. A lot of researchers have focused on detecting emotion \citep{fan_2018a_cnn_facial_expr,xu_2018_emo2vec,winata_2017,winata_2019} and empathy in dialogue systems \citep{bertero_2016_real_sent_rec,chatterjee_2019_a}. \citealp{zhou_2017_emotional_chat_machine} introduced a seq2seq \citep{seq_to_seq_learning_nn} Emotional Chatting Machine in order to generate responses with high emotional context, using emotional embeddings and an internal and external memory mechanism. A GAN-based \citep{goodfellow_2014_gan} framework was also proposed by \citealp{wang_wan_2018_sentigan} that controlled the sentiment of the generated response. \citealp{wu_2019_dual_encoder} also used a dual-decoder to similarly generate emotional responses, given the sentiment.  \citealp{zhou_wang_2017_mojitalk} introduced a Twitter dataset which used  the emojis of the Twitter posts as emotion-labels and they also proposed a seq2seq model to generate emotional responses. \citealp{lubis_2018_eliciting_pos_em} introduced a new dataset and proposed a hierarchical seq2seq response generator for affect-sensitive dialogue generation. \citealp{rashkin_2018_empathetic_dataset} introduced the EmpatheticDialogues dataset and trained the baselines to generate empathetic responses and simultaneously predict the corresponding emotion of the dialogue context. Later, \citealp{lin_2019_moel} introduced the "Mixture of Empathetic Listeners" framework improving the initial baselines. \citealp{santhanam_2019_emotional_nlg} finetuned the GPT2 \citep{Radford_2019_openai_gpt2} model to improve the results further, while \citealp{Shin_2019_happybot} used reinforcement learning for predicting the user's sentiment look-ahead along side with response generation. %In addition, \citealp{Li_2019_acute_eval} introduced ACUTE-EVAl, a novel procedure for  improved pairwise evaluation on full dialogues.
\citealp{lin_2019_caire_empathetic_chatbot}  improved the performance on EmpatheticDialogues by finetuning the GPT2 model with the use of multitask learning, while \citealp{majumder_2020_mime} followed a different approach introducing stochasticity into the emotion mixture and arguing that empathetic responses do not always mirror the emotion of the user. Significant improvements were also made by \citealp{roller_2020_fb_recipes_blender} and \citealp{shuster_2019_dodeca_dialogues} who used multi-task training on multiple dialog tasks, achieving state-of-the-art results.
\\In this work, in order to enforce empathetic response generation we propose a method based on a transformer pretrained language model (T5). Specifically, during finetuning we use three objectives: response language modeling, sentiment understanding and empathy forcing. The sentiment understanding objective is crucial for tracking and acknowledging the emotional state of the conversational partner, while the empathy forcing objective favors empathetic response generation by penalizing responses that have an opposite sentiment of that of the conversational partner.
Our key contribution is the inclusion of the sentiment understanding and empathy forcing auxiliary losses to promote empathetic behavior. The proposed approach, \textit{EmpBot}, \footnote{The implementation will be publicly available after the anonymity period is over}, is on par with state-of-the-art in terms of BLEU score. However, our model produces significantly more fluent and empathetic responses, as indicated by human evaluation results.

% \section{Methodology}\label{sec:Methodology}

\section{Proposed Method}\label{subsec:ProposedMethod}
\begin{figure*}
    \centering
    \includegraphics[width=\textwidth]{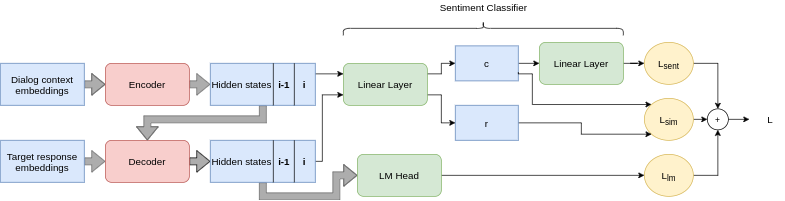}
    \caption{Illustration of the \textit{EmpBot} model. The contextualized sentiment representations $c$ and $r$ are used for calculating the sentiment understanding and the empathy forcing  auxiliary losses.}
    \label{fig:my_label}
\end{figure*}

Our approach is based on the assumption that an empathetic conversational agent should mirror the emotion of the speaker \citep{Carr_2013}. Following this perspective, we introduce \textit{EmpBot}, a model that favors sentiment understanding and  empathetic response generation using the sentiment of each dialogue context.  \textit{EmpBot} is based on  the Unified
Text-to-Text Transformer (T5) \citep{raffel-unified-transformer-T5}, a transformer-based \citep{vaswani-attention-is-all-you-need} pretrained seq2seq network and we extend it with a 2-layer sentiment classifier and auxiliary losses during training, in order to apply sentiment understanding and enforce empathetic response generation. The model is illustrated in Figure \ref{fig:my_label}. 
\noindent\\
\textbf{EmpBot model:} the \textit{EmpBot} model uses the encoded contextualized representations of the dialogue context and the response, to produce the corresponding sentiment representations created by the sentiment classifier, denoted by $c$ and $r$ respectively. It is finetuned on EmpatheticDialogues dataset using three objectives: response language modeling, sentiment understanding and empathy forcing.\\
\textbf{Response language modeling:} to optimize the response language modeling objective we use the contextualized representation of the gold response and we apply language modeling by predicting the reply tokens using the cross-entropy loss. We denote that loss as $L_{lm}$.\\
\textbf{Sentiment understanding:} to optimize the sentiment understanding objective, we pass the contextualized representation of the dialogue context through the 2-layer sentiment classifier and we apply sentiment classification using cross-entropy loss. We denote that loss as $L_{sent}$. In this way, the model learns to predict the sentimental state of the dialogue, and specifically that one of the speaker, using the sentiment labels we created.\\
\textbf{Empathy forcing:} To enforce empathetic behavior, we enhance the model with a cosine similarity embedding loss. Specifically, we use the contextualized sentiment representations obtained from the first layer of the sentiment classifier, both for the dialogue context and the response. The model is penalized, when the sentiment representation of the generated response is different from that of the dialogue context, not only favoring sentiment understanding, but also promoting empathetic response generation. The aforementioned loss is:
    \begin{equation}
    \centering
        L_{sim}(x_1,x_2) = 1-s(x_1,x_2)
    \end{equation}
    where $x_1$, $x_2$ are the contextualized sentiment representations $c$ and $r$ respectively and $s()$ is the cosine similarity function.
    % \end{itemize}
    \noindent
    Our final fine-tuning loss function is the weighted-sum of the aforementioned losses:
    \begin{equation}
    L = L_{lm}+\alpha*L_{sent}+\beta*L_{sim} 
    \end{equation}
    where $\alpha$ and $\beta$ are constants.\footnote{For more details about the hyperparameters tuning see Appendix B}

\section{Experimental Setup}\label{sec:ExpSetup}

\subsection{Dataset}\label{subsec:Dataset}

We conduct our experiments on the EmpatheticDialogues dataset \citep{rashkin_2018_empathetic_dataset}, a dataset consisting of approximately 25k one-on-one open-domain conversations, grounded in a situation and a relevant emotion feeling. For all the experiments, we use the official 8:1:1 train/validation/test split defined by the authors. We group the provided emotions of each dialogue into two groups according to their sentiment polarity. 15 emotions are grouped as positive and 17 as negative.\footnote{For more details about the  split see Appendix A} 
% During testing we do not use the sentiment labels.

\subsection{Models}\label{subsec:Models}
\noindent \textbf{DD MT}: Multitask DodecaDialogue model, proposed in \cite{shuster_2019_dodeca_dialogues}.

\noindent \textbf{DD MT+FT}: Multitask DodecaDialogue model finetuned on EmpatheticDialogues dataset, proposed in \cite{shuster_2019_dodeca_dialogues}.

\noindent \textbf{Baseline}: T5 model finetuned on EmpatheticDialogues for response generation.

\noindent \textbf{EmpBot}: Proposed T5-based model finetuned on EmpatheticDialogues using the proposed loss. Further details for the implementation, the training and testing procedures are provided in Appendix B.

% We experiment with the \textit{EmpBot} and a \textit{baseline} model, evaluating them on EmpatheticDialogues dataset. As a \textit{baseline} model, we use the Unified Text-to-Text Transformer (T5) \citep{raffel-unified-transformer-T5}. The \textit{baseline} is fine-tuned on the EmpatheticDialogues dataset using the response language modeling objective. More specifically,  we use the contextualized representation of the gold response and we apply language modeling by predicting the reply token using the cross-entropy loss. The \textit{EmpBot} is also finetuned on EmpatheticDialogues dataset as mentioned in Section \ref{subsec:ProposedMethod}.

\subsection{Evaluation Protocol}\label{subsec:EvaluationMetrics}
We evaluate our models using both automatic and human evaluation. Although automated metrics can measure both the ability of the model to reproduce the listener's response and the diversity of the responses, they do not always correlate with human judgements of dialogue quality \citep{liu_2016_how_not_to_eval}. Nevertheless, we report both automatic metrics and human evaluation scores. 
\begin{table*}[t!]
\centering
\begin{tabular}{lccc}
\hline
Model &  PPL &  AVG BLEU & Bib \\
\hline
% Vaswani Full Transformer & 21.24  & 6.27 & \citep{rashkin_2018_empathetic_dataset}\\
% Multitask Transformer & 24.07 & 5.42 & \citep{Rashkin_2018_i_know_the_feeling}\\  
% EmoPrepend-1  & 24.30 & 4.36 & \citep{rashkin_2018_empathetic_dataset}\\ 
% TopicPrepend-1  & 25.40 & 4.17 & \citep{rashkin_2018_empathetic_dataset}\\ 
% Ensem-DM & 19.05 & 6.83 & \citep{Rashkin_2018_i_know_the_feeling}\\ 
% Ensem-DM+ & 19.10 & 6.77 & \citep{Rashkin_2018_i_know_the_feeling}\\ 
%CAiRE & 13.32 & 7.03 & \citep{lin_2019_caire_empathetic_chatbot}\\
% GPT2-prepend  & 19.49* & 7.78 & \citep{santhanam_2019_emotional_nlg}\\
%MoEL & - & 2.90 & \citep{lin_2019_moel} \\
%MIME & - & 2.98 & \citep{majumder_2020_mime} \\
%HappyBot & - & 2.32 & \citep{Shin_2019_happybot} \\
BST Generative & $11.48^{\dagger}$ & - & \citep{roller_2020_fb_recipes_blender}\\
DD MT (SOTA) & $11.5$ & $8.4$ & \citep{shuster_2019_dodeca_dialogues}\\
DD MT+FT (SOTA)& $\mathbf{11.4}$ & $8.1$ & \citep{shuster_2019_dodeca_dialogues}\\
Baseline & $12.41$  &  $\mathbf{8.89}$ & -  \\ 
EmpBot  &  $12.37$ & $\mathbf{8.84}$ & - \\\hline
\end{tabular}
\caption{Test performance for automated metrics of the current state-of-the-art approaches, our \textit{baseline} and the \textit{EmpBot} model. Results with $\dagger$ were reported only on the validation set.}
\label{tab:auto_metrics}
\end{table*}

\begin{table*}[t]
\centering
\begin{tabular}{lccc}
\hline
Model &  ${Rel\&Flu}$ &  $Emp_{sent}$ & $Emp_{emo}$ \\
 &  Win | Loss & Win | Loss & Win | Loss \\
\hline
EmpBot vs DD MT+FT & 57.14\% | 42.86\%   &  57.73\% | 42.27\%  &  56.56\% | 43.44\%   \\  
EmpBot  vs Baseline & 63.81\% | 36.19\%  & 65.71\% | 34.29\%   & 59.05\% | 40.95\%   \\
DD MT+FT vs Baseline & 59.05\% | 40.95\%   &  60.95\% | 39.05\%   & 60\% | 40\%     \\
\hline
\end{tabular}
\caption{Results of human A/B testing for each sub-task. All results are statistically significant with p<0.05 using the binomial test.}
\label{tab:pairwise_dod_t5sent}
\end{table*}

\begin{table}[t]
\centering
\begin{adjustbox}{width=\columnwidth,center}
\begin{tabular}{lccc}
\hline
Model Name &  $Relevance$ & $Fluency$ & $Empathy$ \\
\hline
DD MT+FT & 3.42  &  3.96* & 3.33* \\  
EmpBot  &  3.48 & 4.21*  & 3.56* \\\hline
\end{tabular}
\end{adjustbox}
\caption{Human rating test's absolute scores for Relevance, Fluency and Empathy. Results noted with * are statistically significant with p<0.05 using the Mann-Witney U test.}
\label{tab:rating_results}
\end{table}
\noindent
\textbf{Automated metrics}: We report the perplexity (PPL) of the actual (gold) response as in \citealp{Wen_2015,li_diversity_objective_function,li_2016c_persona_conv_model}. Moreover, we  report BLEU scores \citep{papineni_2002_bleu} between the model and the gold response. 

%Finally, we measure the response diversity using the Distinct-1 and Distinct-2 automated metrics introduced by \citep{li_diversity_objective_function}. 
\noindent \textbf{Human evaluation}:
In order to measure the quality  of  the  generated  responses,  we  conduct human evaluation, through an online survey. The human evaluation process is split in two phases. In the first phase, we compare the \textit{EmpBot} model with the current state-of-the-art \textit{DD MT+FT} \citep{shuster_2019_dodeca_dialogues}. Participants were asked to do a pairwise comparison between the generated responses of the aforementioned models according to:
     \textit{Relevance} and \textit{Fluency}  given the dialogue context (denoted by $Rel\&Flu$ in Table \ref{tab:pairwise_dod_t5sent}),
     \textit{Empathy}  given the dialogue context and the speaker's sentiment (denoted by $Emp_{sent}$ in Table \ref{tab:pairwise_dod_t5sent}) and
     \textit{Empathy}  given the dialogue context and the speaker's emotion (denoted by $Emp_{emo}$ in Table \ref{tab:pairwise_dod_t5sent}).
Moreover, participants were also asked to rate each generated response  on the three following aspects: \textit{Relevance}, \textit{Fluency} and \textit{Empathy}, given the dialogue context for each model using a 1-5 Likert scale, where 5 is the best score. So, participants had to complete 3 A/B testing sub-tasks in order to directly compare the models and 2 rating sub-tasks for an indirect comparison. 
%We should note that 7 conversations where randomly sampled from the test set for each sub-task, and totally 49 participants completed the first phase of the survey. \color{black}
\\In the second phase, we compare our \textit{EmpBot} model against our \textit{baseline} and the state-of-the-art \textit{DD MT+FT} model against our \textit{baseline}. In that phase participants were asked to compare the generated responses using the same format of the (3) A/B testing sub-tasks of the first phase.\footnote{For details about the human evaluation  see Appendix C} 
%For each sub-task, 7 conversations where randomly sampled from the test set for each pair of models (2 pairs) and finally 14 conversations where displayed in random order. In this phase, 14 participants took part. 
%\\For testing the statistical significance of the evaluation process, we used the binomial test for the A/B testing sub-tasks as in \citealp{shuster_2019_dodeca_dialogues} and the Mann-Whitney U non-parametric test \citep{Nachar_2008_U_test} for the rating sub-tasks, as it is more robust \citep{Rosenberg2017_MOS} than the t-test in Mean Opinion Scores (MOS) tests.

\section{Results}\label{sec:Results}
Evaluation results and a comparison with other models are presented in Table \ref{tab:auto_metrics}. The human evaluation results are shown in Tables \ref{tab:pairwise_dod_t5sent} and \ref{tab:rating_results}. 
The \textit{DD MT+FT} model still maintains the state-of-the-art performance in perplexity with the \textit{EmpBot} achieving a somewhat lower performance (8.5\% difference). However, we notice that both our \textit{baseline} and the \textit{EmpBot} model outperform the current state-of-the-art model in terms of the average BLEU score by achieving scores as low as 8.89 and 8.84 respectively. Consequently, our empathetic approach (\textit{EmpBot model}) improves the state-of-the-art BLEU score, which was achieved by the \textit{DD MT}, by a difference of 5.2\%. We also notice that our \textit{baseline} performs slightly better on BLEU score metric than the \textit{EmpBot}, but the difference is not  significant. 
\\
However, as the usefulness of the BLEU score has been questioned we turn to human evaluation for a more precise measure of quality. About the human evaluation results, we notice that the \textit{EmpBot} model outperforms both
the \textit{DD MT+FT}
and our \textit{baseline} achieving significantly better results both on A/B and rating tests, as shown in Table \ref{tab:pairwise_dod_t5sent} and \ref{tab:rating_results} respectively. More specifically in Table \ref{tab:rating_results}, we notice a significant difference in \textit{Fluency} and \textit{Empathy} scores, between the \textit{EmpBot} and the \textit {DD MT+FT}, which shows that not only our approach is more empathetic, but the generated responses seem to be more fluent too. About the absolute \textit{Relevance} score, we notice that there is not a significant difference. In addition, we should note that according to the A/B test, shown in Table \ref{tab:pairwise_dod_t5sent} the \textit{DD MT+FT} model seems to perform better than our \textit{baseline} in all sub-tasks. We provide examples of the generated responses in Appendix E.

\section{Conclusions}\label{sec:Conclusions}

In this work we propose \textit{EmpBot}, a T5-based chatbot, augmented with a novel finetuning procedure for generating empathetic dialogue responses.
The proposed loss consists of three parts: an LM loss that produces valid textual responses, a sentiment classification loss that introduces emotional awareness to the model and an empathy forcing loss that ensures that the responses are emotionally relevant.
We evaluate \textit{EmpBot} using standard evaluation metrics, i.e. perplexity and BLEU score, achieving state-of-the-art results.
Our human evaluation results indicate that \textit{EmpBot} produces more fluent and empathetic responses, when compared with both the baseline and the state-of-the-art models.
In the future we want to extend the proposed method for other architectures, and explore %more fine-grained emotion and 
more empathy forcing losses using raw emotion values instead of sentiment polarities.

\bibliographystyle{acl_natbib}
\bibliography{custom}
\clearpage
\appendix

\section{Dataset Seperation Split Details}\label{sec:data_sep_details}
We split the 32 provided emotion annotations according to their sentiment polarity,  as  illustrated in Table \ref{tab:emotions}.
\begin{table}[htb]
\centering
\small
\begin{tabular}{|l|l|}
\hline
Positive & Negative \\
\hline
 \begin{tabular}{l}surprised, excited, \\proud, grateful,\\ impressed, hopeful,\\ confident, joyful,\\ content, caring,\\ trusting, faithful,\\ prepared, sentimental,\\ anticipating 
\end{tabular} & \begin{tabular}{l}
     angry, sad,\\ annoyed,lonely,\\ afraid, terrified,\\ guilty,disgusted,\\ furious, anxious,\\ nostalgic, disappointed,\\ jealous, devastated,\\ embarrassed, ashamed,\\ apprehensive 
\end{tabular}\\
\hline

\end{tabular}
\caption{Separation split of 32 emotions based on their valence.}
\label{tab:emotions}
\end{table}

\section{Implementation \& Training Details}\label{sec:train_details}
We use the T5-base model from the HuggingFace library having 12 layers, 768 hidden-states, 3072 feed-forward hidden-states and 12 heads. We also use a 300-d dimensional space for the sentiment representations obtained from the 2-layer classifier. Finally, the \textit{baseline} and the \textit{EmpBot} model have $\sim222$M and $\sim223$M parameters respectively.\\
During training, we set $\alpha$ to $0.4$ and $\beta$ to $0.4$. After experimenting with various empirically selected value pairs for the parameters $\alpha$ and $\beta$, we found that the selected values yield the slightly best PPL for the validation set. We use the Adam optimizer, setting the learning rate equal to $2\mathrm{e}{-5}$ and the weight decay to $1\mathrm{e}{-6}$. We also use a batch size of 4. All hyperparameters were manually tuned and the set with the best validation perplexity was chosen. All models were trained in a single Tesla K80 GPU provided by Google Colab. \\
During inference time, we use top-p (nucleus) sampling method \citep{Hotzman_2019_topp} with top-k filtering \citep{Fan_2018_topk}, by setting threshold probability $p$ equal to 0.9 and $topk$ to 10. We also add length penalty equal to 0.6 and we set the maximum length of the generated response to be equal to 40.

\section{Human Evaluation Details}\label{sec:eval_details}
% In the first phase, 7 conversations where randomly sampled from the test set for each sub-task (both for A/B and rating tests), and totally 49 participants completed the first phase of the survey. 
% \\
% In the second phase, for each sub-task of A/B testing, 7 conversations where randomly sampled from the test set for each pair of models (\textit{EmpBot} vs \textit{Baseline} and \textit{DD MT+FT} vs \textit{Baseline}) and finally 14 conversations where displayed in random order. In this phase, 15 participants took part. 
 The human evaluation study was completed by proficient English speakers, volunteers responding to a corresponding request we posted at our university’s and research institute’s communication channels. All tests were blind and the participants could not tell which model the various dialogue responses to be evaluated were coming from. For the A/B testing, participants were asked to select the best-generated response (according to Relevance and Fluency, Empathy given the emotion of the context, and Empathy given the sentiment of the context - 3 sub-tasks ). For the rating tests, participants were asked to rate (1-5 Leikert scale) each model independently (2 sub-tasks) in terms of Empathy, Relevance and Fluency. The following clarifications were given for each metric: \enquote{Relevance evaluates whether the generated response is on-topic with the dialogue context}, \enquote{Fluency measures the grammatical correctness and readability of the generated response} and \enquote{Empathy measures whether the generated response shows the understanding of the speaker’s feelings}. 
\\
During our the evaluation study each user was presented with 7 conversations. These conversations were randomly sampled from the whole test set (2547 conversations). Therefore the participants were presented with 343 (as 49 participants took part in the first phase) and 105 (as 15 participants took part in the second phase) unique conversations, in total, during the first and second phases of the study respectively.
\\
For testing the statistical significance of the evaluation process, we used the binomial test for the A/B testing sub-tasks as in \citealp{shuster_2019_dodeca_dialogues} and the Mann-Whitney U non-parametric test \citep{Nachar_2008_U_test} for the rating sub-tasks, as it is more robust \citep{Rosenberg2017_MOS} than the t-test in Mean Opinion Scores (MOS) tests.

\section{Additional Results}\label{sec:add_results}
A full comparison,  based on automatic evaluation, between the \textit{baseline}, the \textit{EmpBot} model and other existing approaches is presented in Table \ref{tab:add_auto_metrics}.
\begin{table*}[t!]
\centering
\begin{tabular}{lccc}
\hline
Model &  PPL &  AVG BLEU & Bib \\
\hline
Vaswani Full Transformer & $21.24$  & $6.27$ & \citep{rashkin_2018_empathetic_dataset}\\
% Multitask Transformer & $24.07$ & $5.42$ & \citep{Rashkin_2018_i_know_the_feeling}\\  
EmoPrepend-1  & $24.30$ & $4.36$ & \citep{rashkin_2018_empathetic_dataset}\\ 
TopicPrepend-1  & $25.40$ & $4.17$ & \citep{rashkin_2018_empathetic_dataset}\\ 
% Ensem-DM & $19.05$ & $6.83$ & \citep{Rashkin_2018_i_know_the_feeling}\\ 
% Ensem-DM+ & $19.10$ & $6.77$ & \citep{Rashkin_2018_i_know_the_feeling}\\ 
CAiRE & $13.32$ & $7.03$ & \citep{lin_2019_caire_empathetic_chatbot}\\
GPT2-prepend  & $19.49^{\dagger}$ & $7.78$ & \citep{santhanam_2019_emotional_nlg}\\
MoEL & - & $2.90$ & \citep{lin_2019_moel} \\
MIME & - & $2.98$ & \citep{majumder_2020_mime} \\
HappyBot & - & $2.32$ & \citep{Shin_2019_happybot} \\
BST Generative & $11.48^{\dagger}$ & - & \citep{roller_2020_fb_recipes_blender}\\
DD MT (SOTA) & $11.5$ & $8.4$ & \citep{shuster_2019_dodeca_dialogues}\\
DD MT+FT (SOTA)& $\mathbf{11.4}$ & $8.1$ & \citep{shuster_2019_dodeca_dialogues}\\
Baseline & $12.41$  &  $\mathbf{8.89}$ & -  \\ 
EmpBot  &  $12.37$ & $\mathbf{8.84}$ & - \\\hline
\end{tabular}
\caption{Test performance for automated metrics of the existing approaches, our \textit{baseline} and the \textit{EmpBot} model. Automated metrics noted with $\dagger$ were reported only on validation set.}
\label{tab:add_auto_metrics}
\end{table*}

\section{Example Outputs}\label{sec:example_outputs}
We provide some examples of the generated responses for both the \textit{EmpBot}, the \textit{baseline} and the \textit{DD MT+FT} models in Table \ref{tab:examples}.

\begin{table*}[t!]
\centering
\begin{tabularx}{0.9\textwidth}{l}
\toprule
\textbf{Example 1} \\
\hline
Speaker: I felt so let down by a person who promised me a project -- but then pulled it \\away at the last minute in a way that was very personal and reactionary.\\\hline
Target: That's terrible. People need to learn to commit to their promises.\\
Emotion: disappointed\\ \hline
\textit{EmpBot}: That's not good. I'm sorry you felt that way.\\
\textit{baseline}: That's terrible. What was the project?\\
\textit{DD MT+FT}: what was the project about ?\\
\end{tabularx}

\begin{tabularx}{0.9\textwidth}{l}
\toprule
\textbf{Example 2} \\
\hline
Speaker: I had to give a presentation in front of my entire University class last semester. I \\hate presenting things.\\\hline
Target: oh, that is nerve wracking. What did you try to do to cope?\\
Emotion: anxious\\ \hline
\textit{EmpBot}: That's a real shame. I'm sure you did a great job.\\
\textit{baseline}: I'm sorry to hear that.\\
\textit{DD MT+FT}: how did it go ?\\
\end{tabularx}

\begin{tabularx}{0.9\textwidth}{l}
\toprule
\textbf{Example 3} \\
\hline
Speaker: I am going to my local amusement park tomorrow and feeling a certain way about\\ it! \\\hline
Target: Which way are you feeling? Are you a thrill seeker?\\
Emotion: excited\\ \hline
\textit{EmpBot}: That sounds like fun! I love going to the amusement park!\\
\textit{baseline}: I'm sorry to hear that.\\
\textit{DD MT+FT}: what amusement park are you going to ?\\
\end{tabularx}

\begin{tabularx}{0.9\textwidth}{l}
\toprule
\textbf{Example 4} \\
\hline
Speaker: I felt very isolated when we first moved to Hong Kong -- I had a young baby and \\ my husband pretty much needed to start traveling for business right away!\\\hline
Target: Wow, I can't imagine how you felt living in Hong Kong. That would take some\\ serious adapting.\\
Emotion: lonely\\ \hline
\textit{EmpBot}: I'm sorry to hear that. Is he still in Hong Kong?\\
\textit{baseline}: That's a bummer. I'm sorry to hear that.\\
\textit{DD MT+FT}: i ' m sorry to hear that . did you have a lot of fun ?\\
\end{tabularx}
\caption{Examples of the generated responses for each model, along with the target response and the corresponding emotion of the speaker.}
\label{tab:examples}
\end{table*}

\end{document}